\documentclass[a4paper]{article}

\usepackage{INTERSPEECH2021}

\usepackage[hyphens]{url}
\usepackage{
    footmisc,
    hyperref,
}

\hypersetup{
    colorlinks=false,
    linkcolor=blue,
    citecolor=blue,
    urlcolor=blue,
}

\usepackage[dvipsnames]{xcolor}

\title{AlloST: Low-resource Speech Translation without Source Transcription}
\name{Yao-Fei Cheng$^{1}$, Hung-Shin Lee$^{2}$, and Hsin-Min Wang}
\address{Institute of Information Science, Academia Sinica, Taiwan}
\email{$^{1}$freddy@iis.sinica.edu.tw, $^{2    }$hungshinlee@gmail.com}

\begin{document}

\maketitle

\begin{abstract}

The end-to-end architecture has made promising progress in speech translation (ST). However, the ST task is still challenging under low-resource conditions. Most ST models have shown unsatisfactory results, especially in the absence of word information from the source speech utterance. In this study, we survey methods to improve ST performance without using source transcription, and propose a learning framework that utilizes a language-independent universal phone recognizer. The framework is based on an attention-based sequence-to-sequence model, where the encoder generates the phonetic embeddings and phone-aware acoustic representations, and the decoder controls the fusion of the two embedding streams to produce the target token sequence. In addition to investigating different fusion strategies, we explore the specific usage of byte pair encoding (BPE), which compresses a phone sequence into a syllable-like segmented sequence. Due to the conversion of symbols, a segmented sequence represents not only pronunciation but also language-dependent information lacking in phones. Experiments conducted on the Fisher Spanish-English and Taigi-Mandarin drama corpora show that our method outperforms the conformer-based baseline, and the performance is close to that of the existing best method using source transcription.
\end{abstract}

\section{Introduction}

Communication is the basic requirement of human daily life. However, more than 7,000 different languages are spoken in the world. Due to the growing demand for international activities, it is very important to break the communication barriers between different spoken languages. Unlike text-to-text machine translation (MT), which takes text as input to generate text in another language, the input of speech translation (ST) is just a speech signal stream. Traditionally, the task of ST is divided into two connected processes: automatic speech recognition (ASR) and MT \cite{F.Casacuberta}. The advantage of this cascading approach is that it can make the best use of text and speech resources. However, this approach is prone to error propagation, i.e., word errors generated by ASR may lead to incorrect translations. Not only the source speech and target translation, the training process of this approach also requires the transcription of the source speech, which imposes restrictions on all unwritten languages and some low-resource languages. Therefore, directly translating the source speech to the target text is essential and has been extensively developed in recent research \cite{Berard2016,Weiss2017,Gulati2020}. Compared with the cascading approach, the end-to-end approach avoids the problem of error-propagation, reduces the inference time, and does not necessarily require transcription of the source speech for training. 

When the transcription of the source speech is available, the ST model that implements the end-to-end method of joint training of ASR and MT components can effectively translate the given source speech \cite{Anastasopoulos2018}. Source speech recognition becomes a sub-task, which helps to obtain more phonetically informative embeddings of the source speech. Moreover, the jointly trained model can provide good initial model parameters for subsequent fine-tuning of the ST-only task \cite{Liu2019,Chuang2020,Wang2020,CWang2020}. On the other hand, in the cascading approach, it has been proved that the phone sequence of the source speech derived from the word sequence output of a well-trained ASR system has better performance than the word sequence in ST \cite{Salesky2019,Salesky2020}. In other words, phone features are a better substitute for the textual resources of a language. Salesky and Black \cite{Salesky2020} further proved that phone features are very effective for both cascading and end-to-end ST approaches under \textit{low-resource} conditions. The results leave us certain things to imagine: Can we obtain reliable phonetic information of a speech utterance of an unwritten language? This is one of our motivations for this study.

As far as we know, the end-to-end ST approach achieves promising results because there are a large amount of resources available for both speech and text. Although these methods work well under high-resource conditions, they still fail in low-resource and unwritten languages. Due to the difficulty of collecting the language-dependent lexicon and phone inventory, any method that uses text resources is unrealistic for low-resource and unwritten languages. For these languages, text resources usually do not exist or are in poor condition. For example, over 3,000 languages have not yet developed a written form, but are still in use. Therefore, how to directly translate and document low-resource unwritten languages becomes a problem we are trying to solve \cite{Adda2016,Muller2017,Boito2017}. In accordance with the actual situation, we follow the restrictions to simulate a real-world low-resource unwritten language in our ST task: 1) any lexicons, phone inventories, pre-trained ASR and MT models of the source language cannot be used; 2) only approximately 40 hours of unlabeled source speech are available; 3) external language-independent resources can be used.

Therefore, in this study, we propose a hybrid learning framework that does not require any source language-dependent resources, such as word or phone transcription of speech, lexicons, and phone inventories. In \cite{Li2020}, Li \textit{et al.} proposed Allosaurus, a neural system for extracting a universal phone sequence from a speech utterance. Allosaurus was built using the PHOIBLE dataset, a curated phone dataset containing more than 2,000 languages. Owing to its comprehensiveness and validity for the phones of various languages, we use it as our phone recognizer for extracting phone features, regardless of whether the decoded phone sequence is represented according to the phonetic specification of the source language. Furthermore, we propose two methods to integrate phone information into the acoustic features and the sequence-to-sequence model for ST. Our experiments are conducted on two low-resource corpora, namely the 40-hour subset of the Fisher Spanish-English corpus and the 30-hour Taigi-Mandarin drama corpus. The results show that our method outperforms the conformer-based end-to-end baseline \cite{Gulati2020}, and the performance is very close to that of the existing best approach using source transcription. This work, named AlloST, is open-sourced\footnote{\url{https://github.com/jamfly/AlloST}}. The main contribution of this study is two-fold. First, we confirm the potential of low-resource ST that does not require source transcription; to the best of our knowledge, all existing cascading methods require it. Second, as an improvement of the end-to-end approach, we prove that the segmentation of phone information is very helpful to boost the performance of ST.

\begin{figure}[t]
\centering
\includegraphics[width=0.45\textwidth]{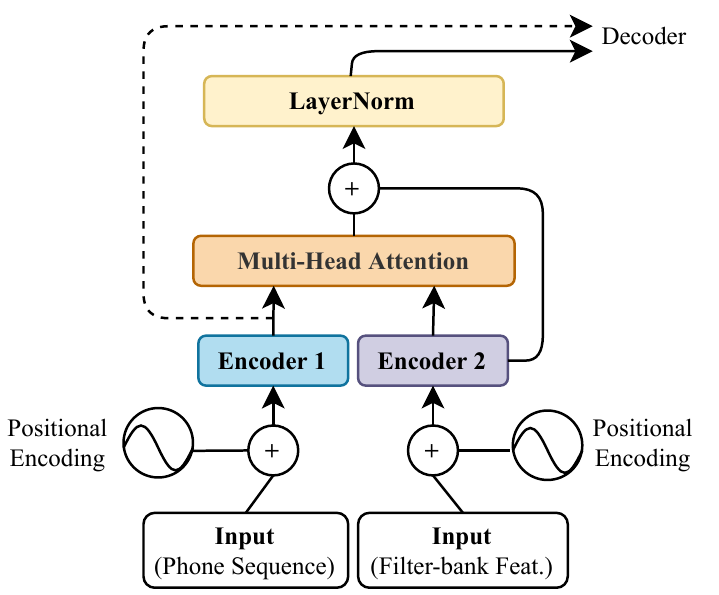}
\vspace{-5pt}
\caption{The architecture of the encoder in the proposed model.}
\label{fig:encoder}
\vspace{-15pt}
\end{figure}

\section{Proposed Model: AlloST}

\subsection{Allosaurus}

The traditional phone recognition method either recognizes speech to the international phonetic symbols (e.g., IPA) of all languages or the phones of a specific language. Allosarus, a language-independent phone recognizer \cite{Li2020}, aims to increase the phone recognition accuracy under low-resource conditions. Allosarus leverages knowledge from other languages to predict the phone inventory of an unknown language. Unwritten languages may not have a pre-defined phone inventory; however, spoken units are language-independent, and they are just symbols of pronunciation. Allosaurus meets our need because it can leverage the phone inventories across 2,000 different languages to recognize speech in any language to language-independent or language-dependent symbols.

Allosaurus has two execution modes: 1) producing language specific phone sequences for known languages; 2) producing the IPA sequence for any language (the default setting) with the phone inventory consisting of approximately 230 phones. In our experiments, we adopt the default setting because we assume that the source language is unknown. Because Allosaurus's training data contain a Spanish corpus, it produces more accurate phone recognition results for Spanish than Taigi. This difference is reflected in our experimental results.


\subsection{Methods}

\subsubsection{Phone features}

We assume that phone labels can be used not only to express pronunciation, but also to convey a certain level of semantics. Based on the idea that the combination of phone labels can be regarded as a morpheme of the language, we adopt BPE to segment a phone sequence to find possible morpheme units \cite{Rico2016}. Originally, BPE was used to solve the problem of out-of-vocabulary (OOV). However, it can also efficiently capture the relationship between pronunciation and semantics within the corpus. We use trainable embeddings with positional encoding to transfer semantic information from raw speech to high-level presentations, and then fuse them with acoustic features. We will introduce the strategies of integrating these presentations into existing models in the following sections.

\begin{figure}[t]
\centering
\includegraphics[width=0.45\textwidth]{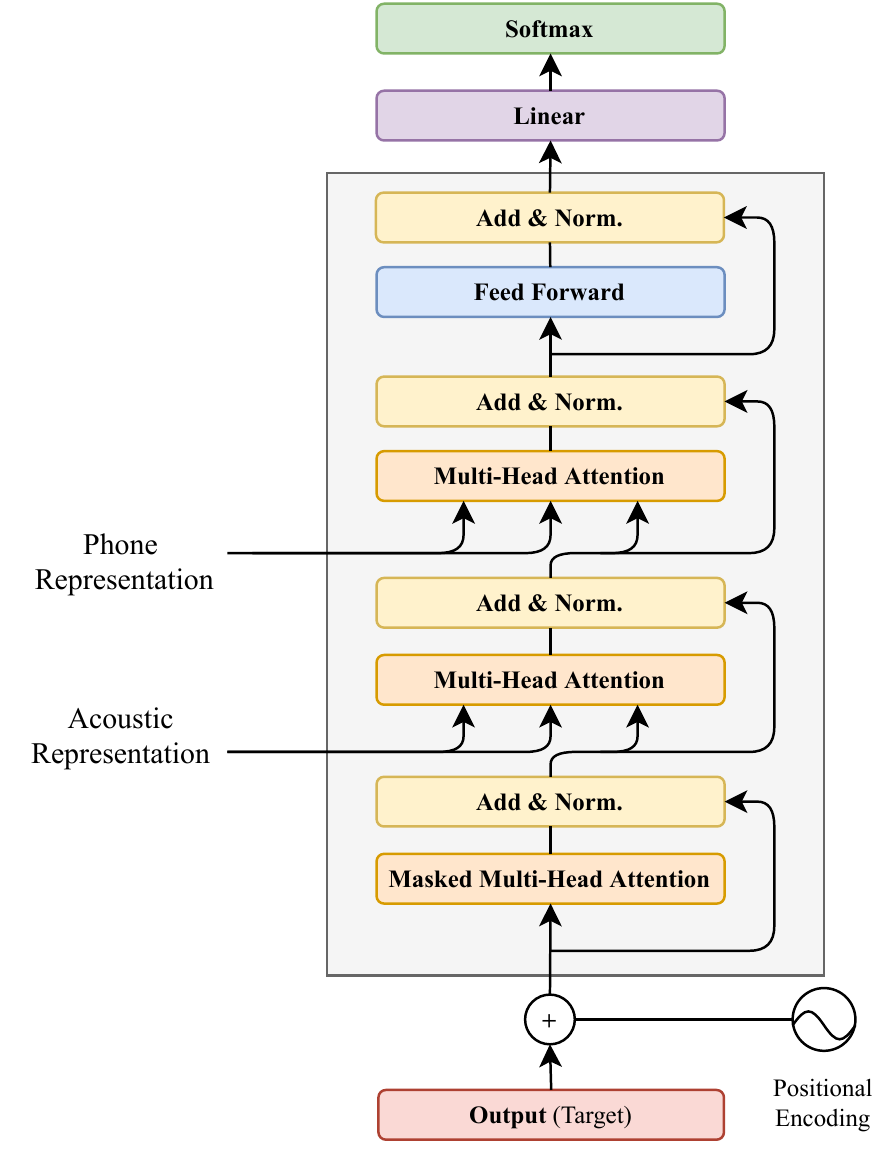}
\vspace{-5pt}
\caption{The architecture of the decoder in the proposed model.}
\label{fig:decoder}
\vspace{-15pt}
\end{figure}

\subsubsection{Encoder fusion}

As shown in Figure \ref{fig:encoder}, we use two separate conformer encoders \cite{Gulati2020} to model phone and acoustic features. Given a source speech utterance, Encoder 1 takes the phone sequence as input while Encoder 2 takes the  acoustic feature (filter-bank) sequence as input.

Salesky \textit{et al.} \cite{Salesky2019,Salesky2020} proposed two methods to integrate phone features and acoustic features. The first one is to concatenate the trainable phone embeddings and acoustic embeddings, and the second one is to use the phone boundaries to segment and compress the acoustic embeddings. However, both methods are not applicable here because Allosaurus's phone sequence output does not contain time information. To solve the problem, we propose to use multi-head attention to learn the alignment between phone features and acoustic features \cite{Vaswani2017}.

\subsubsection{Decoder fusion}

Inspired by recent research on using a second decoder to do post-editing \cite{Le2020,Lee2019,Sung2019,Sperber2019,Shin2018}, we use a similar structure to achieve the goal of proofreading. As shown in Figure \ref{fig:decoder}, we use the basic setting of the transformer decoder \cite{Vaswani2017}, but add an additional stacked multi-head attention layer after the original multi-head attention layer to deal with the phone embedding of the source speech. We believe that the phone representation can provide detailed and supplementary information about the source speech, and when it is fused with the resulting representation after the first attention layer that deals with the acoustic representation, it can correct ``mishearing'' errors. The decoder generates one token at a time, and then feeds that token back to the decoder to generate the next token until the stop symbol appears. 
In addition to the stacked multi-head attention, we also tried to use the gate (type (b)) in \cite{Libovicky2018} to fuse phone and acoustic features; however, it did not improve the quality of translation.

\begin{table}[t]
\caption{BLEU scores on the Fisher Spanish-English corpus with BPE 1k. ($\star$) indicates that source transcription is used, and ($\ddagger$) means that additional Wikipedia resources are used.}
\vspace{-5pt}
\setlength{\tabcolsep}{4pt}
\label{tab:fisher_results}
\footnotesize
\centering
\begin{tabular}{lccc}
\toprule
\bf{Method} & \bf{fisher-dev} & \bf{fisher-dev2} & \bf{fisher-test} \\
\midrule\midrule
Conformer \cite{Gulati2020} & 20.84 & 21.49 & 21.06 \\
$\star$ Conformer w/ MTL \cite{Inaguma2020} & 31.21 & \bf{32.59}& 31.47 \\
$\star$ phone Segment \cite{Salesky2019} & 21.00 & N/A & 19.80 \\
$\star$ phone Concat. \cite{Salesky2020} & \bf{34.50} & N/A & \bf{33.00} \\ 
$\ddagger$ Word Embedding \cite{Chuang2020} & 23.54 & N/A & 21.72 \\
\midrule
Encoder Fusion & \bf{29.52} & 30.01 & 29.01 \\
Decoder Fusion & 26.32 & 27.14 & 26.19 \\
Encoder + Decoder Fusion & 28.95 & \bf{30.18} & \bf{29.49} \\
\bottomrule
\end{tabular}
\vspace{-10pt}
\end{table}

\section{Experimental Setup}

\subsection{Datasets}

\subsubsection{Fisher Spanish-English corpus}

The Fisher Spanish-English speech translation corpus (LDC2014T23) is a widely used corpus for ST research \cite{Post2013}. The corpus contains 160 hours of Spanish telephone speech, including 138k utterances. All 160 hours of speech have corresponding transcription and English translation. We selected 40 hours of speech from the corpus to conduct experiments that simulate low-resource conditions. We used the standard dev, dev2, and test sets to evaluate our methods, each set containing approximately 4k utterances.

\subsubsection{Taigi-Mandarin drama corpus}

Taigi, also known as Taiwanese Minnan, is a variant of Southern Min spoken in Taiwan. There are two main types of writing systems, ``Pinyin'' and ``Han-Luo''. However, most people are not familiar with any Taigi writing system. According to the current situation, Taigi can be regarded as an unwritten language. Chen \textit{et al.} collected 9 TV series of Taigi dramas with Mandarin Chinese subtitles. It contains about 870 hours of speech, of which 30 hours of speech are manually labeled with Mandarin Chinese translation \cite{Chen2020}. We used the 30-hour manually labeled speech as the training set, and used approximately 6,000 speech utterances from different dramas as the test set.

\subsection{Pre-processing}

We followed the ESPnet-ST recipe to extract 80 log-mel filterbank coefficients and 3-dimensional pitch features using the Kaldi toolkit as the acoustic feature \cite{Inaguma2020,Povey2011}. Speed perturbation with the factor of 3 was applied to the training source speech \cite{Ko2015}. The punctuation marks in the translation were removed. The training pairs with target translation longer than 400 characters or source speech longer than 3,000 frames were discarded. Allosaurus was applied to generate the phone sequence from source speech. BPE was applied to both the source phone sequence and the target translation. 
For the two target languages, the vocabulary size of BPE was fixed at 1k. For the phone sequences of the two source languages, the vocabulary size ranged from 1k to 64k.
BPE-Dropout with dropout-rate 0.1 was only applied to the phone sequences \cite{Provilov2020}.

\subsection{Implementation Details}

The conformer architecture implemented in ESPnet-ST was used as the baseline system \cite{Inaguma2020}. The encoder consists of 12 layers, and the decoder consists of 6 layers. We used 4-head self-attention with the dimension of 256 \cite{Vaswani2017}. The dimension of feed-forward was set to 2,048, and label-smoothing and dropout were set to 0.1. The Adam optimizer \cite{Kingma2015} and Noam learning rate schedule \cite{Vaswani2017} with default settings were used. The proposed model was implemented on top of the baseline. Different from the baseline system, dropout and label-smoothing were set to 0.3. The additional encoder (Encoder 1 in Figure \ref{fig:encoder}) consists of 6 layers. An additional stacked 4-head attention layer was used in the decoder.

For evaluation, the model parameters corresponding to the 5 best checkpoints based on the performance of the validation set were averaged. For the Fisher Spanish-English corpus, we reported the 4-reference BLEU score \cite{Post2013} because there are 4 reference translations for each source utterance.

\section{Experimental Results}

\begin{table}[t]
\caption{BLEU scores on the Taigi-Mandarin drama corpus with BPE 1k.}
\vspace{-5pt}
\label{tab:taigi_results}
\footnotesize
\centering
\begin{tabular}{ccc}
\toprule
\bf{Method} & \bf{taigi-test} & \bf{BLEU $\triangle$} \\ 
\midrule\midrule
Conformer & 18.01 & \\
\midrule
Encoder Fusion & 22.69 & +4.68 \\
Decoder Fusion & \bf{22.94} & \bf{+4.93} \\
Encoder + Decoder Fusion & 22.58 & +4.57 \\
\bottomrule
\end{tabular}
\vspace{-15pt}
\end{table}

\begin{table*}[t]
\caption{The results of different ST models for two examples in the Fisher Spanish-English corpus. The text in green stands for correct translations in this example, and there are no missing words in other methods. The text in blue stands for correct translations in this example, but there are missing words in other methods. The text in red stands for wrong translations. }
\vspace{-5pt}
\setlength{\tabcolsep}{4pt}
\label{tab:examples}
\footnotesize
\centering
\begin{tabular}{ccc}
\toprule
\bf{Method} & \bf{Example 1} & \bf{Example 2} \\
\midrule\midrule

Ground Truth & you buy music or \textcolor{ForestGreen}{have} many \textcolor{blue}{cds} & now \textcolor{ForestGreen}{that's} going to be in a very \textcolor{blue}{little time} my friend \\
\midrule
Conformer & that eh do you buy music or do you \textcolor{ForestGreen}{have} lots & \textcolor{red}{they're doing it in puerto rico} \\
Conformer w/ MTL & that uh do you buy music or \textcolor{red}{has} a lot of \textcolor{blue}{cds} & \textcolor{ForestGreen}{it's} going to be a \textcolor{blue}{little time} \textcolor{red}{ago} \\
\midrule
Encoder + Decoder Fusion w/o BPE & \textcolor{red}{what} eh you buy music or do you \textcolor{ForestGreen}{have} \textcolor{red}{much difficult} & \textcolor{ForestGreen}{it's} going to be a \textcolor{red}{long time} for me \\
Encoder + Decoder Fusion w/ BPE 1k & that eh do you buy music or \textcolor{ForestGreen}{have} many \textcolor{blue}{cds} & \textcolor{ForestGreen}{it's} going to be a \textcolor{blue}{little time} for me \\
\bottomrule
\end{tabular}
\vspace{-10pt}
\end{table*}

\begin{table}[t]
\caption{Impact of BPE on the BLEU scores on the Fisher Spanish-English corpus.}
\vspace{-5pt}
\label{tab:bpe_impact}
\footnotesize
\centering
\begin{tabular}{ccccc}
\toprule
\bf{Method} & \bf{fisher-dev} & \bf{fisher-dev2} & \bf{fisher-test} & BLEU $\triangle$ \\
\midrule\midrule
w/o BPE & 26.52 & 26.92 & 26.12 \\
\midrule
BPE 1k & 28.95 & 30.18 & 29.49 & +3.33 \\
BPE 10k & 29.26 & 29.90 & 29.37 & +3.25 \\
BPE 16k & 28.89 & 30.18 & 29.73 & +3.61 \\
BPE 32k & 29.19 & \bf{31.05} & \bf{30.30} & \bf{+4.18} \\
BPE 48k & \bf{29.37} & 30.26 & 29.70 & +3.58 \\
BPE 64k & 29.44 & 30.73 & 29.12 & +3.00 \\
\bottomrule
\end{tabular}
\vspace{-15pt}
\end{table}

\subsection{Main Results}

The BLEU scores evaluated on the Fisher Spanish-English corpus are shown in Table \ref{tab:fisher_results}. The proposed models (``Encoder Fusion'', ``Decoder Fusion'' and ``Encoder + Decder Fusion'') are compared with several models including models that leverage extra resources. ``Conformer'' was trained in an end-to-end manner without using any text resources of the source language \cite{Gulati2020}. ``Conformer w/ MTL'' \cite{Inaguma2020} was trained with multi-task learning of hybrid CTC/Attention ASR and MT models \cite{S.Watanabe2017}, where the transcription of Spanish speech was used to train the ASR model. ``phone Segment'' \cite{Salesky2019} and ``phone Concat.'' \cite{Salesky2020} were trained with additional Spanish text resources (LDC96L16). ``Word Embedding'' \cite{Chuang2020} used FastText that was pre-trained on Wikipedia word embeddings.

From Table \ref{tab:fisher_results}, we can see that our models consistently outperform the baseline conformer model (``Conformer'') on all the three test sets, e.g., the BLEU score was improved from 21.06\% to 29.49\% on the fisher-test set. Our models also outperform ``phone Segment'', which used additional Spanish text resources, and ``Word Embedding'', which used pre-trained word embeddings as extra information. We believe that compared with word embeddings, phone features are closer to acoustic features, and it is rather easy to combine phone features with acoustic features. Obviously, our models are slightly worse than ``phone Concat.'' and ``Conformer w/ MTL''. However, ``phone Concat.'' used segmented phone features derived from ASR, while ``Conformer w/ MTL'' used the transcription of the source speech to jointly train the ASR and MT models. The results show that segmented phone features are more effective.

\subsection{Ablation Studies}

In order to further understand which strategy is more suitable for fusing phone features and acoustic features, we conducted an ablation study. From Table \ref{tab:fisher_results}, the results show that fusing phone features and acoustic features in both Encoder and Decoder at the same time is generally more effective than fusing phone features and acoustic features in Encoder or Decoder alone. Fusing the phone features with the acoustic features and the target tokens that have been translated in Decoder alone are not as effective as fusing phone features and acoustic features in Encoder alone. We guess this is because the phone features are similar to the acoustic features in Encoder. Therefore, fusing the phone features and the acoustic features in Encoder is easier to find the alignment, so as to obtain better results.

The results in Table \ref{tab:fisher_results} indicate that the effectiveness of using the phone features in Decoder to correct the translation is limited because the source phone features and the target tokens are quite different. However, this may only happen in the case when the source language and the target language are very different. The second task of this study is ST for the Taigi-Mandarin pair. The results are shown in Table \ref{tab:taigi_results}. Taigi and Mandarin share many vocabularies and have similar grammar. We can think of this task as translation between two similar languages. From Table \ref{tab:taigi_results}, we can see that ``Decoder Fusion'' is more effective than ``Encoder Fusion'', and is even more effective than the combination of the two. The results show that using the Taigi phone representations to correct the translated Mandarin text is an effective way to improve the quality of translation.

\subsection{Examples}

Table \ref{tab:examples} shows the translation results of different models for two examples in the fisher-test set. The first row is ground truth, the second and third rows are the translations of ``Conformer'' and ``Conformer w/ MTL''. In the first example, the translation of ``Conformer'' missed the word ``cds'', and our proposed model ``Encoder + Decoder Fusion w/ BPE 1k'' could accurately translate this word. ``Conformer w/ MTL'' caused grammatical errors (``have'' vs. ``has''), and our proposed models ``Encoder + Decoder Fusion w/o BPE'' and ``Encoder + Decoder Fusion w/ BPE 1k'' did not make this error.

In the second example, ``Conformer'' failed to translate the entire sentence. ``Conformer w/ MTL'' incorrectly translated ``ago'', and because of this error, the meaning of the sentence is completely different. Our proposed model ``Encoder + Decoder Fusion w/ BPE 1k'' obviously output a better translation.

\subsection{BPE vs. Raw Phone Sequence}

Motivated by previous work that segmented speech into a phone sequence according to phone change boundaries \cite{Salesky2019}, we applied BPE with BPE-Dropout of dropout-rate 0.1 \cite{Rico2016,Provilov2020} to reduce the phone sequence length. As the BPE unit inventory increases from 230 (the size of IPA) to 1k $\sim$ 64k, the phone sequence length can be reduced by about 30\% $\sim$ 60\%. We believe that combining BPE-based phone features will enable the model to have a deeper understanding of sentences.

In order to analyze the effect of BPE segmentation, we conducted experiments using different sizes of BPE vocabulary. As shown in Table \ref{tab:bpe_impact}, BPE-based phone features consistently outperform raw phone features. As shown in Table \ref{tab:examples}, the model that fused the raw phone features with the acoustic features (cf. ``Encoder + Decoder Fusion w/o BPE'') mistranslated ``many cds'' to ``much difficult'' in example 1 and ``little time'' to ``long time'' in example 2, but the model that fused the BPE-based phone features with the acoustic features (cf. ``Encoder + Decoder Fusion w/ BPE 1k'') did not make such errors. One possible reason is that fusing the raw phone features with the acoustic features will make the model pay too much attention to the pronunciation and ignore the meaning of the sentence, thereby resulting in inaccurate translation.

\section{Conclusions}

In this study, we propose a framework called AlloST, which shows that language-independent phone features generated by a universal phone recognizer can improve low-resource speech translation tasks. We also show that fusing the phone features with the acoustic features in the encoder, decoder or both can generally improve the performance, but the best setting depends on the language pairs. Generally, BPE-based phone sequence segmentation is of great benefit to performance. Our conformer-based model only uses an additional universal phone recognizer (i.e., Allosaurus). But it is superior or comparable to several conformer-based models that use more other resources.

\section{Acknowledgments}

This work was supported in part by MOST-Taiwan under Grant: 110-2634-F-008-004.

\newpage
\bibliographystyle{IEEEtran}

\bibliography{references.bib}
\end{document}